\pdfoutput=1

\documentclass[11pt]{article}

\usepackage[review]{EMNLP2023}

\usepackage{times}
\usepackage{latexsym}
\usepackage{multirow}
\usepackage{booktabs}

\usepackage{colortbl}
\usepackage[T1]{fontenc}
\usepackage{graphicx}

\usepackage[utf8]{inputenc}

\usepackage{microtype}

\usepackage{inconsolata}
\usepackage{tikz}
\usepackage{collcell}
\usepackage{enumitem}

\newcommand*\mycirc[1]{%
\begin{tikzpicture}[baseline=(C.base)]
\node[draw,circle,inner sep=1pt,minimum size=.5ex](C) {#1};
\end{tikzpicture}}
\usepackage{hyperref}
\usepackage{multirow}
\usepackage{rotating}

\title{GD-COMET: A Geo-Diverse Commonsense Inference Model}

\author{Mehar Bhatia \and Vered Shwartz \\
University of British Columbia\\
Vector Institute for AI\\
{\tt \{meharb23,  vshwartz\}@cs.ubc.ca
}}

\newcommand{\model}{\textsc{gd-comet}}

\begin{document}
\maketitle
\begin{abstract}
With the increasing integration of AI into everyday life, it's becoming crucial to design AI systems that serve users from diverse backgrounds by making them culturally aware. In this paper, we present \model{}, a geo-diverse version of the COMET commonsense inference model. \model{} goes beyond Western commonsense knowledge and is capable of generating inferences pertaining to a broad range of cultures. We demonstrate the effectiveness of \model{} through a comprehensive human evaluation across 5 diverse cultures, as well as extrinsic evaluation on a geo-diverse task. The evaluation shows that \model{} captures and generates culturally nuanced commonsense knowledge, demonstrating its potential to benefit NLP applications across the board and contribute to making NLP more inclusive.

\end{abstract}

\section{Introduction}
\label{sec:intro}
Culture plays a significant role in shaping an individual's worldviews, beliefs, behaviours, and communication styles \cite{spradley1987culture}. A considerable portion of what is commonly referred to as commonsense knowledge is not universal but rather culture-specific, including social norms, values, traditions, and more. An example of cultural differences is greetings, which may involve a handshake in Western cultures, bowing in some Asian cultures, a `namaste' gesture in India, or `wai' in Thailand. 

\begin{figure}
    \centering
    \includegraphics[width=.43\textwidth]{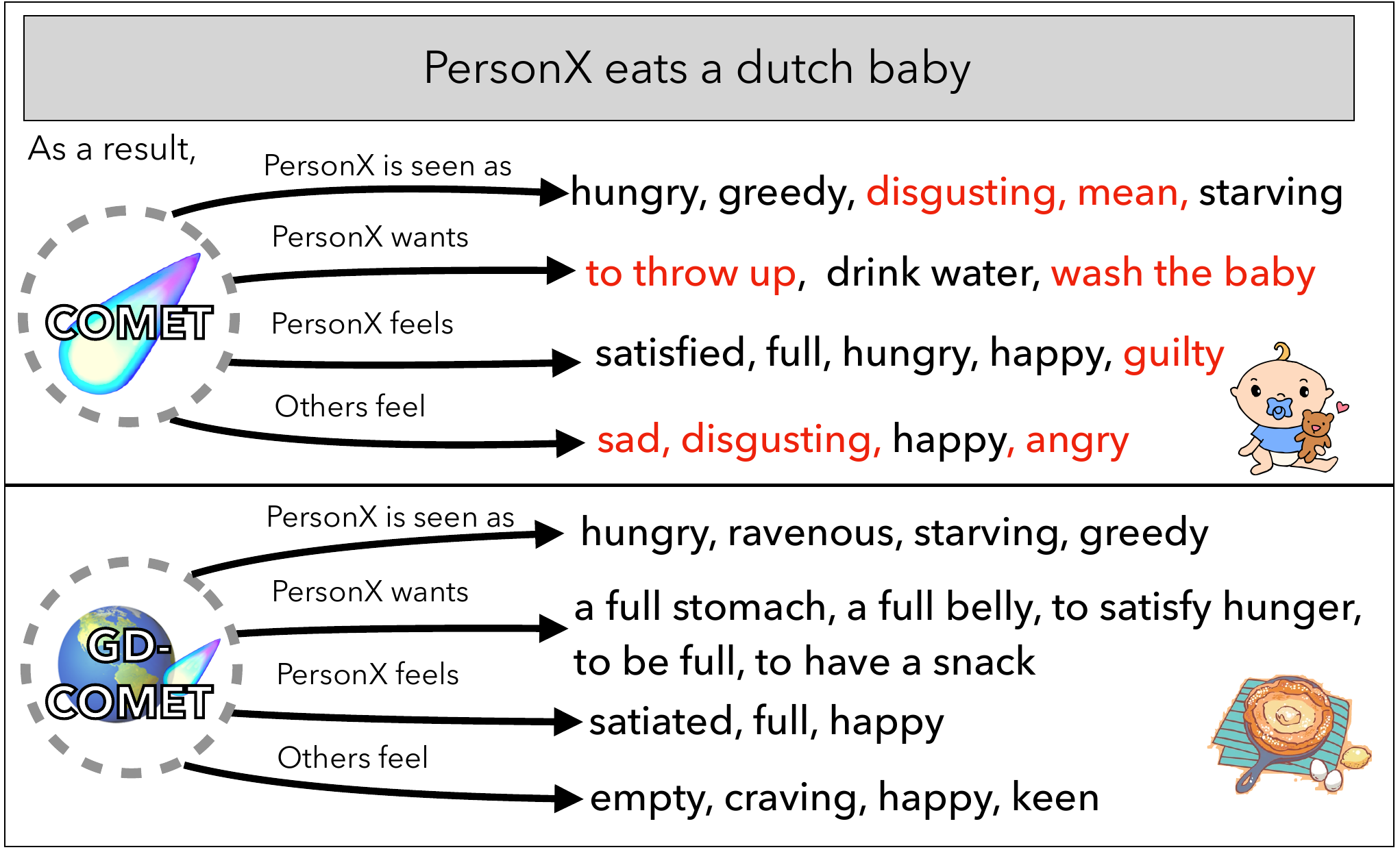}
    \vspace{-5pt}
    \caption{Inferences from COMET and \model{} for the sentence ``PersonX eats a dutch baby'', demonstrating lack of culture awareness in COMET.}
    \label{fig:dutch_baby}
    \vspace{-10pt}
\end{figure}

With AI systems becoming increasingly ubiquitous in society, it is imperative to go beyond the Western cultural perspective \cite{hershcovich-etal-2022-challenges}. Lack of cultural awareness may lead to models perpetuating stereotypes and reinforcing societal inequalities \cite{hutchinson-etal-2020-social,ross-etal-2021-measuring,sogaard-2022-ban}, impeding their effectiveness for users from non-Western countries. 

In this paper, we focus on a popular model for commonsense reasoning, COMET \cite{bosselut-etal-2019-comet}, which is based on an English language model (LM) and further trained on commonsense inferences collected from North American crowdsource workers \cite{sap2019atomic}. Consequently, the model exhibits a certain bias towards the North American cultural perspective. As evidenced by Fig.~\ref{fig:dutch_baby}, COMET displays limited familiarity with the concept of a German pancake, erroneously interpreting the term ``dutch baby'' in a literal sense.

We identify a need for more inclusive commonsense reasoning models and propose \model{}: \textbf{G}eo-\textbf{D}iverse COMET. As demonstrated in Fig~\ref{fig:dutch_baby}, \model{} gained the culturally relevant knowledge to interpret ``dutch baby'' as a legitimate dish. 

\model{} is similarly based on an English LM but is trained on a knowledge base of cultural knowledge \cite{10.1145/3543507.3583535} prior to training on COMET's original training data. This simple approach is effective, as judged by both human evaluations as well as extrinsic evaluation on a geo-diverse task \cite{yin-etal-2021-broaden}. \model{} can potentially benefit many downstream NLP applications where the user population is diverse.\footnote{Code available at \href{https://github.com/meharbhatia/GD-COMET}{github.com/meharbhatia/GD-COMET}}

\section{Background}
\label{sec:background}
\subsection{Commonsense Inference Models}
\label{sec:bg:commonsense_models}

Many NLP tasks require reasoning beyond what is explicitly stated in the text. People fill in those gaps with their commonsense knowledge. NLP models attempt to do the same by leveraging commonsense knowledge bases (KBs) such as ConceptNet \cite{speer2017conceptnet} and ATOMIC \cite{sap2019atomic}. To achieve better coverage, knowledge models such as COMET \cite{bosselut-etal-2019-comet} are based on pre-trained LMs 
and further fine-tuned on KBs, enabling contextually-relevant inferences along the KB's dimensions for new contexts. 

COMET's hybrid approach proved useful for various tasks \cite[e.g.,][]{chakrabarty-etal-2020-r,
ammanabrolu2021automated,ravi2023vlc}. Subsequent versions of COMET have been developed to draw inferences from paragraphs \cite{gabriel2021paragraph}, images \cite{park2020visualcomet}, and complex sentences \cite{ravi2023comet}. Further improvements include obtaining additional training data through crowdsourcing \cite{hwang2021comet} or generating synthetic data from LMs \cite{west-etal-2022-symbolic}. 

COMET and its successors assume the universality of commonsense knowledge, yet much of this knowledge may differ among cultures, in traditions \cite[e.g., duration of a wedding ceremony;][]{acharya2021towards}, foods \cite[e.g., what counts as breakfast food;][]{speer2017conceptnet}, social norms, and more. 

\subsection{Culture-Aware NLP}
\label{sec:bg:culture-aware-nlp}

While multilingual NLP is a popular topic, culture-aware NLP is under-explored. It is crucial for language technologies to not only serve speakers of a wide variety of languages but also acknowledge that users come from diverse cultures \cite{hershcovich-etal-2022-challenges}. 
Cultural norms and pragmatic aspects differ across speakers from different cultures \cite{zhou-etal-2023-cross}. Nevertheless, English LMs primarily reflect a North-American lens due to training on web data with a US user bias \cite{cao-etal-2023-assessing}.

Current work in culture-aware NLP addresses various aspects. One line of work focuses on cultural stereotypes and biases, and ways to measure and mitigate them \cite[e.g.,][]{hutchinson-etal-2020-social,ross-etal-2021-measuring,sogaard-2022-ban}. Another line of work analyzes the differences in culture-specific commonsense knowledge, including 
relational knowledge \cite{yin-etal-2022-geomlama}, grounding of time expressions \cite{shwartz-2022-good}, food-related customs \cite{palta-rudinger-2023-fork} and social values \cite{lin-etal-2021-common,arora-etal-2023-probing}. At the same time, there have been efforts to develop benchmarks \cite{yin-etal-2021-broaden,liu-etal-2021-visually}, and adapt models to new cultures \cite{zhou-etal-2023-cross,yin2023givl}. Finally, there are several recent cultural KBs such as StereoKG \cite{deshpande-etal-2022-stereokg}, Quasimodo \cite{romero2019commonsense}, and CANDLE \cite{10.1145/3543507.3583535}. CANDLE, which we use in this work, is the most comprehensive among them, containing 1.1M assertions in English about 386 cultures (e.g. ``A Dutch baby is a German pancake that is baked instead of cooked on the stove top''). CANDLE assertions were extracted from a large web corpus and clustered into \emph{5 facets of culture}: food, drinks, clothing, rituals, and traditions. 

\section{\model{}}
\label{sec:model}
We present \model{}, a geo-diverse version of COMET. The goal of \model{} is to generate high-quality commonsense inferences for concepts and events pertaining to both Western and non-Western cultures. Rather than collecting a large-scale geo-diverse dataset in the style of ATOMIC, we split the training into two phases: (1) training the underlying LM on geo-diverse data; (2) continue training on the large-scale original COMET training data. This is motivated by \newcite{bosselut-etal-2019-comet} that showed that implicit commonsense knowledge from underlying LM's pre-training transfers to COMET. We similarly hypothesize that encoding geo-diverse data into the underlying LM prior to training on COMET data will transfer this knowledge to \model{}.

\paragraph{Geo-Diverse Training (GD-BART).} We pick 770,000 assertions from CANDLE with a combined score greater than 0.5. This threshold selects highly distinctive assertions specific and relevant to the specific region. We fine-tune BART-Large, the underlying LM of the latest COMET model \cite{hwang2021comet}, on this data, using BART's original pre-training objectives (token masking, token deletion, text infilling and sentence permutation). We save the model checkpoint with the lowest validation loss after training for 50 epochs on two NVIDIA A40 GPUs.


\paragraph{COMET Training.} We proceed to fine-tuning GD-BART on the large-scale ATOMIC-2020 dataset, using the same training method and hyper-parameters as \newcite{hwang2021comet}. Appendix~\ref{appendix:comet-relations} lists the 34 COMET relations used in this paper.

\section{Intrinsic Evaluation}
\label{sec:intrinsic}

\newcommand*{\MinNumber}{1.75}%
\newcommand*{\MaxNumber}{3.7}%

\newcommand{\ApplyGradient}[1]{%
    \ifdim #1 pt > \MinNumber pt
        \pgfmathsetmacro{\PercentColor}{max(min(100.0*(#1 - \MinNumber)/(\MaxNumber-\MinNumber),100.0),0.00)} %
        \colorbox{red!\PercentColor!white}{#1}
    \else
        \pgfmathsetmacro{\PercentColor}{max(min(100.0*(#1 - \MinNumber)/(\MaxNumber-\MinNumber),100.0),0.00)} %
        \colorbox{white!\PercentColor!white}{#1}
        \fi
}

\newcolumntype{R}{>{\collectcell\ApplyGradient}c<{\endcollectcell}}

\newcommand*{\MinNum}{1.75}%
\newcommand*{\MaxNum}{3.7}%

\newcommand{\ApplyGrad}[1]{%
    \ifdim #1 pt > \MinNum pt
        \pgfmathsetmacro{\PercentColor}{max(min(100.0*(#1 - \MinNum)/(\MaxNum-\MinNum),100.0),0.00)} %
        \colorbox{orange!\PercentColor!white}{#1}
    \else
        \pgfmathsetmacro{\PercentColor}{max(min(100.0*(#1 - \MinNumber)/(\MaxNum-\MinNum),100.0),0.00)} %
        \colorbox{white!\PercentColor!white}{#1}
        \fi
}

\newcolumntype{L}{>{\collectcell\ApplyGrad}c<{\endcollectcell}}

\begin{table}[t]
    \centering
    \small
    \begin{tabular}{ll*{4}{L}}
         \toprule
         & & \multicolumn{1}{c}{\mycirc{1}} & \multicolumn{1}{c}{\mycirc{2}} & \multicolumn{1}{c}{\mycirc{3}} & \multicolumn{1}{c}{Average $\kappa$}\\ \midrule
         \multirow{5}{*}{\begin{turn}{90}COMET\end{turn}} & \textbf{India} & 2.32 & 2.16 & 2.65 & 0.71\\
         & \textbf{S Korea} & 1.93 & 1.86 & 2.32 &  0.67\\
         & \textbf{Nigeria} & 1.97 & 1.98 & 2.27 &  0.61\\
         & \textbf{Iran} & 2.09 & 2.31 & 2.42 & 0.63\\
         & \textbf{Indonesia} & 2.28 & 2.36 & 2.55 &  0.66\\
    \end{tabular}
    \begin{tabular}{ll*4{R}}
         \toprule
         & & \multicolumn{1}{c}{\mycirc{1}} & \multicolumn{1}{c}{\mycirc{2}} & \multicolumn{1}{c}{\mycirc{3}} & \multicolumn{1}{c}{Average $\kappa$}\\ \midrule
         \multirow{5}{*}{\begin{turn}{90}\model{}\end{turn}} & \textbf{India} & 2.62 & 2.54 & 2.73 & 0.74\\
         & \textbf{S Korea} & 2.13 & 1.92 & 2.35 & 0.65  \\
         & \textbf{Nigeria} & 2.25 & 1.92 & 2.35 & 0.59  \\
         & \textbf{Iran} & 2.27 & 2.38 & 2.58 & 0.76\\
         & \textbf{Indonesia} & 2.43 & 2.46 & 2.58 & 0.77 \\
         \bottomrule
    \end{tabular}
    \caption{Evaluation of COMET and \model{} inferences, judged by annotators from the respective cultures.}
    \label{tab:intrinsic_results} 
    \end{table}

To evaluate the quality of \model{}, we construct a set of input sentences pertaining to 5 diverse cultures (Table~\ref{tab:intrinsic_results}). We sample 5 concepts for each facet and use facet-specific templates (Appendix~\ref{appendix:category_template}) to create 20 sentences for each culture. For each of COMET and \model{}, we use beam search to generate 5 inferences for each of the 34 dimensions and convert them to natural language statements using relation-specific templates based on prior work \cite{bosselut-etal-2019-comet}. The correctness of both inferences were judged by 10 graduate students, two students from each of the respective cultures. Annotators were asked to grade inferences along the following criteria on scale of 0 (worst) to 3 (best):

\begin{enumerate}[leftmargin=*,itemsep=0em,label=\protect\mycirc{\arabic*},topsep=0pt]
    \item \textbf{Cultural Relevance:} The inference is factually accurate and reflects the values, customs, traditions, and societal norms associated with the given culture. 
    \item \textbf{Stereotype Avoidance:} The inference does not perpetuate stereotypes about the culture.
    \item \textbf{Linguistic Accuracy:} The inference is grammatical, and the vocabulary and idiomatic expressions are appropriate in that culture.
\end{enumerate}

The annotations yielded a substantial inter-annotator agreement with $\kappa$ = 0.656 for COMET and 0.702 for \model{}, measured with average Cohen's Kappa \cite{cohen1960coefficient} across cultures.

\paragraph{Results.} Table~\ref{tab:intrinsic_results} reveals that \model{} consistently outperforms the standard COMET model. Specifically, \model{} excels in generating culturally aligned inferences across chosen diverse cultures, and is more likely than COMET to avoid biased assumptions. However, there is still room for improvement for South Korea and Nigeria.

\section{Extrinsic Evaluation}
\label{sec:extrinsic}
Traditional benchmarks often fall short in testing models' knowledge and comprehension of diverse cultural contexts. To show \model{}'s utility for downstream tasks, we evaluate on a multimodal task, GD-VCR (Sec~\ref{sec:extrinsic:data}). We develop a model inspired by VLC-BERT \cite{ravi2023vlc} that generates inferences and incorporates them into a vision and language (V\&L) model (Sec~\ref{sec:extrinsic:model}). We show that \model{} improves the performance on GD-VCR upon an array of baselines (Sec~\ref{sec:extrinsic:results}) and demonstrate the inferences contributing to the performance gains (Sec~\ref{sec:qualitative:analysis}). 

\subsection{Dataset}
\label{sec:extrinsic:data}

\begin{table*}[t]
    \centering
    \small
    \begin{tabular}{l|c|cc|c|ccc}
        \toprule
       \multirow{2}{*}{\textbf{Datasets}} & \multirow{2}{*}{\textbf{Human}} & \multirow{2}{*}{\textbf{VisualBERT*}} & \multirow{2}{*}{\textbf{ViLBERT*}} &\multirow{2}{*}{\textbf{VL-BERT}} &\multicolumn{3}{c}{\textbf{VLC-BERT w/}}\\
       & & & & & \textbf{GD-BART} & \textbf{COMET} & \textbf{\model{}} \\
        \midrule
        \textbf{GD-VCR}              & 88.84 & 53.27 & 58.47 & 58.63  & 52.69   & 59.59   & \textbf{63.51}\\
        \midrule
        \textbf{ $\circ$ West}       & 91.23 & 65.82 & 64.37 & 65.27  &  57.69  &  66.78  & \textbf{69.93}\\
        \textbf{ $\circ$ South Asia} & 92.98 & 52.04 & 62.9  & 64.92  &  54.35  &  64.25  & \textbf{68.17}\\
        \textbf{ $\circ$ Africa}     & 87.93 & 51.85 & 62.04 & 58.17  &  51.87  &  57.71  & \textbf{64.81}\\
        \textbf{ $\circ$ East Asia}  & 83.05 & 45.39 & 46.45 & 47.88  &  41.87  &  49.64  & \textbf{53.07}\\
        \bottomrule
    \end{tabular}
    \caption{\small{Accuracy (\%) of the different models on the subset of each region in GD-VCR. 
    We report the average across 3 runs (see Appendix~\ref{appendix:performance} for the results of individual seeds). Results marked with $*$ have been reported in \newcite{yin-etal-2021-broaden}.}}
    \label{tab:results}
\end{table*}

Visual Commonsense Reasoning \cite[VCR;][]{zellers2019recognition} is a benchmark for testing V\&L models' ability to understand and reason beyond a visual scene. Each example consists of an image extracted from movies or TV series and a multiple-choice question about the actions or people depicted in the image. This dataset focuses solely on Western, primarily North American movies. 

Geo-Diverse Visual Commonsense Reasoning dataset \cite[GD-VCR;][]{yin-etal-2021-broaden} follows the same setup of VCR but extends to diverse regions. This evaluation-only dataset includes 328 images from movies and TV series in East Asian, South Asian, African and Western countries (See Appendix \ref{appendix:gdvcr_stats}). We follow the original setup and train our model on VCR before testing on GD-VCR.

\subsection{Model (VLC-BERT with \model{})}
\label{sec:extrinsic:model}

\begin{figure}[!ht]
    \centering
    \includegraphics[width=\linewidth]{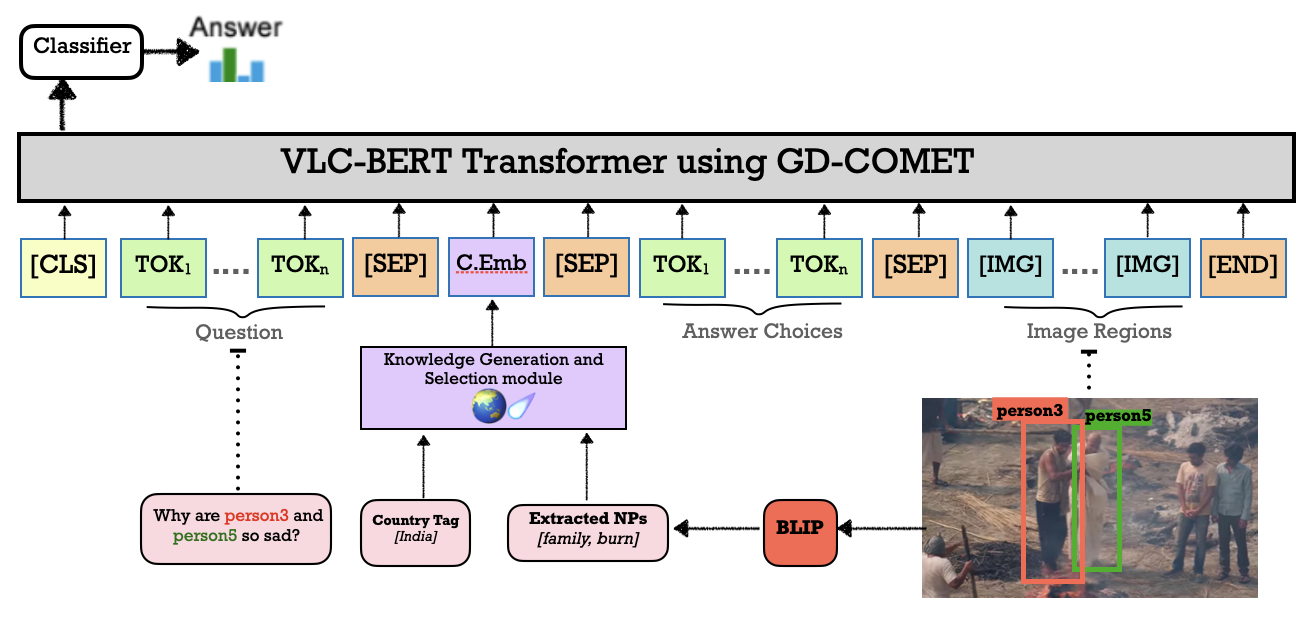}
    \caption{A model using \model{} for GD-VCR.}
    \label{fig:eval-setup-gdcomet}
\end{figure}

We take inspiration from VLC-BERT \cite{ravi2023vlc}, that incorporated COMET inferences into VL-BERT \cite{su2019vl}. Instead, we integrate GD-COMET as a source of contextualized cultural commonsense knowledge for GD-VCR. Figure~\ref{fig:eval-setup-gdcomet} illustrates the model. We describe below VLC-BERT and where our model deviates from it.  

\paragraph{Knowledge Generation and Selection.} VLC-BERT uses the question and the object tags as input to COMET. Instead of object tags, we generate an image caption using BLIP \cite{li2023blip} and extract noun phrases from the caption using SpaCy \cite{honnibal2020spacy}. We found that the noun phrases provide a more detailed description of the depicted activities within the image (e.g. ``\texttt{family, burn}'' in Fig.~\ref{fig:eval-setup-gdcomet}). We additionally append a country tag to the input. During training on VCR, we use the tag ``\texttt{North America}'', the primary source of movies in the dataset. For the images in GD-VCR, we extracted country tags from Wikipedia. 

We use beam search to generate five inferences for each of the 34 dimensions. To select the most relevant inferences, we convert the inferences to natural language sentences using relation-specific templates and select the inferences that are the most similar to the question using SBERT embeddings \cite{reimers-gurevych-2019-sentence}. 

\paragraph{Overall Architecture.} The generic input to VL-BERT for VCR is \texttt{<question, answer tokens, image regions>}. Following \newcite{ravi2023vlc}, 
we embed each inference with SBERT and summarize them into a single token with a weighted average based on learned attention scores. Finally, we feed the output of the \texttt{[CLS]} token into a classifier to predict a score for each answer choice. We train the model using binary cross-entropy loss for 20 epochs on 4 NVIDIA RTX6000 GPUs.

\begin{figure*}[t]
     \centering
     \includegraphics[width=\textwidth]{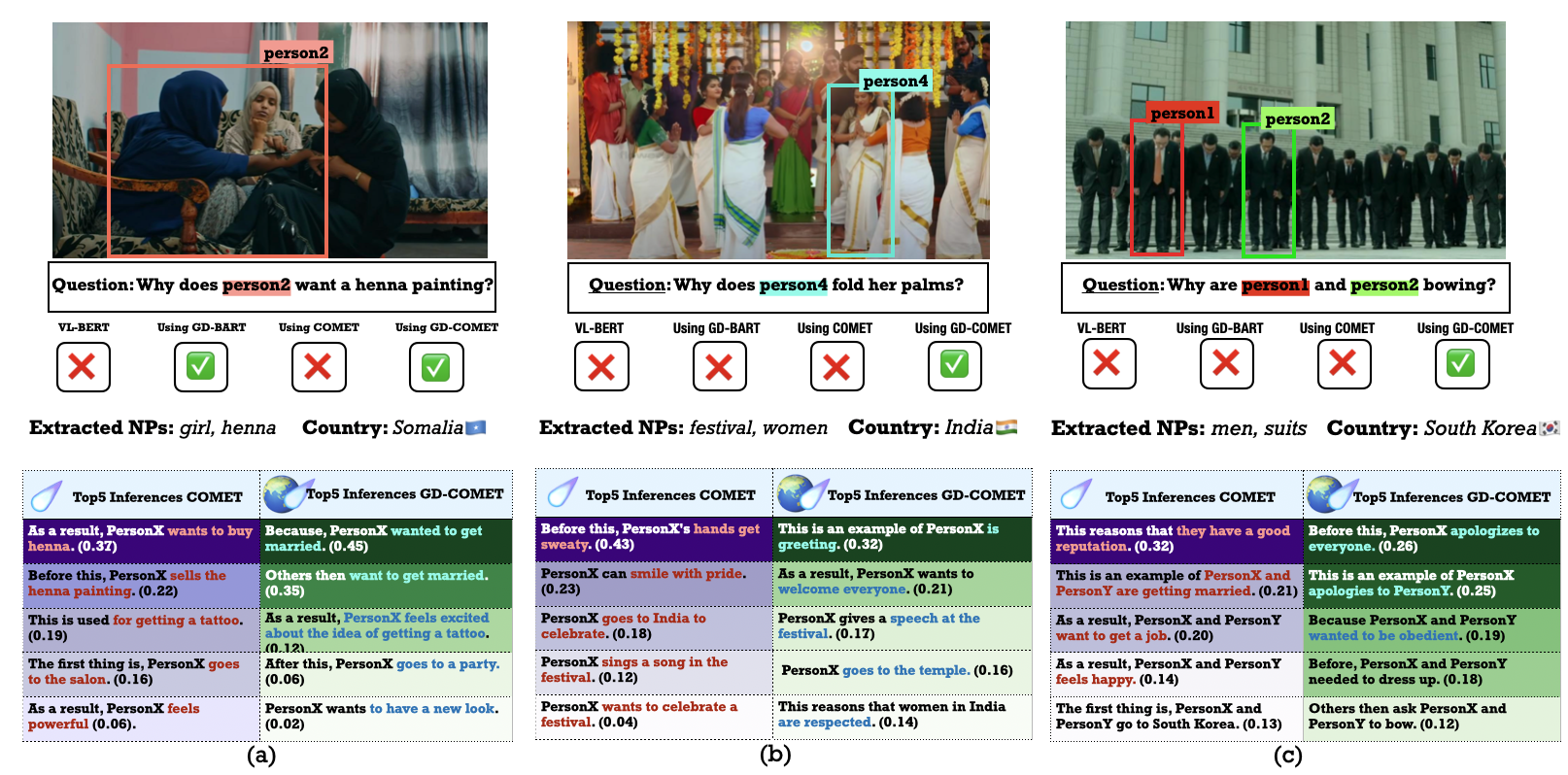}
     \caption{Attention analysis of commonsense inferences generated by COMET and \model{} for testing samples in GD-VCR.} 
     \label{fig:attention-analysis}
 \end{figure*}

\subsection{Results}
\label{sec:extrinsic:results}

Table~\ref{tab:results} compares our model's performance on GD-VCR with baselines that: (i) do not make use of commonsense knowledge (VL-BERT); (ii) generate inferences using GD-BART; and (iii) use COMET (VLC-BERT w/COMET). Note that the same signals (i.e., country tag and noun phrases) were used for the GD-BART and COMET baselines. We also include prior results reported using VisualBERT and ViLBERT for completeness.  

VLC-BERT w/COMET modestly improves upon VL-BERT across most regions, with an overall improvement of 1.2 points in accuracy. This suggests that COMET provides some commonsense inferences that are universal. 
Conversely, \model{} shows a substantial improvement of nearly 5 points over VL-BERT and 4 points over VLC-BERT w/COMET. This highlights the effectiveness of incorporating \model{} for downstream tasks that require culture-specific knowledge across diverse regions. Furthermore, GD-BART performs less effectively than other methods, underscoring the importance of training on structured knowledge to generate contextually relevant responses.

\subsection{Qualitative Analysis}
\label{sec:qualitative:analysis}

Figures~\ref{fig:attention-analysis} presents several GD-VCR instances along with the models' predictions, and the inferences generated by COMET and \model{} for them. In Figure~\ref{fig:attention-analysis}a, \model{} accurately associates a girl wearing henna in Somalia with marriage. In Figure~\ref{fig:attention-analysis}b, it understands that folding palms during an Indian festival signifies a greeting or welcome. Finally, in Figure~\ref{fig:attention-analysis}c, it recognizes that bowing in South Korea is a gesture of apology, leading to VLC-BERT w/ \model{} to be the only model that provides a correct answer. In contrast, COMET's inferences for this example are generic and irrelevant. These examples highlight \model{}'s effectiveness in identifying the cultural context and dynamically generating culturally-relevant commonsense inferences across ATOMIC's relations.

\section{Conclusion}
\label{sec:conclusion}
This work challenges the current notion of universally applicable commonsense knowledge by introducing \model{}, a geo-diverse variant of COMET. \model{}  can generate culturally-nuanced commonsense inferences for a broad range of cultures. Our comprehensive evaluation confirms the effectiveness of \model{} in incorporating and leveraging cultural cues. 
We view our work as a step towards developing more inclusive and culturally-aware AI systems.

\section*{Limitations}
While \model{} represents a significant advancement in incorporating cultural commonsense knowledge into AI models, a few limitations need to be acknowledged.

First, the availability of comprehensive, high-quality data remains a challenge in training culturally-aware models. While resources like CANDLE provide a step forward in curating diverse cultural knowledge, it is essential to note that merely capturing the existence of concepts within a culture is insufficient. Future efforts should aim to collect data that reflects the presence of certain concepts and encompasses how people perceive and interpret those concepts within their specific cultural contexts. This would require extensive data collection efforts that go beyond surface-level understanding, and delve into the nuances of cultural perspectives. 

A second limitation is the availability of suitable benchmarks for testing models' knowledge and understanding of cultural variations. In particular, two such tasks, GD-VCR and MarVL \cite{liu-etal-2021-visually}, focus on vision and language, while \newcite{10.1145/3543507.3583535} proposes a cultural knowledge quiz. We hope to see more language-only datasets developed to go beyond testing models on knowledge about concepts from diverse cultures to understanding cultural nuances.

\section*{Ethics Statement}
Despite being designed to be more culturally inclusive, \model{} runs the risk of unintentionally perpetuating biases present in CANDLE data. In particular, CANDLE might misrepresent cultures with stereotypes or underrepresent cultures. Addressing these concerns requires proactive measures such as identifying biases using methods such as \newcite{mehrabi-etal-2021-lawyers} and mitigating them through filtering and additional data collection.  

Additionally, the size of evaluation benchmarks means they don't always account for cultural variations within the same region. For example, GD-VCR images in the African region are concentrated in East Africa. Similarly, addressing this issue would require additional annotation efforts.

\section*{Acknowledgement}
This work was funded, in part, by the Vector Institute for AI, Canada CIFAR AI Chairs program, an NSERC discovery grant, and a research gift from AI2. Finally, we sincerely thank Sahithya Ravi, Aditya Chinchure, Ward Pennink and Jan Zimny for valuable feedback and discussions.

\bibliography{anthology,custom}
\bibliographystyle{acl_natbib}
\appendix

\section{COMET Relations}
\label{appendix:comet-relations}
Table~\ref{tab:relations-list} lists COMET relations used in this work. 
\begin{table}[!ht]
\centering
\scriptsize
\tt
\begin{tabular}{lll}
\toprule
AtLocation & CapableOf & isBefore \\
Causes & CausesDesire & isFilledBy \\
CreatedBy & Desires & oEffect \\
HasPrerequisite & HasFirstSubevent & oReact \\
HasA & HasProperty & oWant \\
InstanceOf & IsA & xAttr \\
LocatedNear & MadeOf & xEffect \\
MadeUpOf & MotivatedByGoal & xIntent \\
ObjectUse & PartOf & xNeed \\
ReceivesAction & SymbolOf & xReact \\
UsedFor & isAfter & xReason \\
xWant \\
\bottomrule
\end{tabular}
\caption{COMET relations used in this work.}
\label{tab:relations-list}
\end{table}

\section{Facet Templates}
\label{appendix:category_template}

Table~\ref{tab:category-templates} presents the templates used for creating input sentences to \model{} for each concept associated with a cultural facet. 

\begin{table}[!ht]
\centering
\scriptsize
\tt
\begin{tabular}{ll}
\toprule
\textbf{clothing} & ``PersonX wears [concept] in [country]'' \\
\textbf{food} & ``PersonX eats [concept] in [country]'' \\
\textbf{drink} & ``PersonX drinks [concept] in [country]'' \\
\textbf{festival} & ``PersonX celebrates [concept] in [country]'' \\
\bottomrule
\end{tabular}
\caption{Templates used to create input sentences for \model{} for each CANDLE facet.}
\label{tab:category-templates}
\end{table}

\section{VCR and GD-VCR Statistics}
\label{appendix:gdvcr_stats}

Table~\ref{tab:GD-VCR dataset} displays the statistics of the VCR and GD-VCR datasets. The bottom half shows the number of images for each region in GD-VCR. 

\begin{table}[h]
    \scriptsize
    \centering
    \begin{tabular}{lrrrr}
        \toprule
        & \textbf{\# Images} & \textbf{\# QA Pairs} & \textbf{avg Q} & \textbf{avg A}\\
        & & & \textbf{length} & \textbf{length}\\
        \midrule
        \textbf{VCR (dev)} & 9929 & 26534 & 6.77 & 7.67 \\
        \midrule
        \textbf{GD-VCR} & 328 & 886 & 7.38 & 7.68 \\
        \midrule
        \textbf{West} & 100 & 275 & 7.36 & 7.19 \\
        \textbf{East Asia} & 101 & 282 & 7.59 & 7.59 \\
        \textbf{South Asia} & 87 & 221 & 6.85 & 8.00 \\
        \textbf{Africa} & 40 & 108 & 7.98 & 8.54 \\
        \bottomrule
    \end{tabular}
    \caption{\small{Statistics of the VCR and GD-VCR benchmarks.}}
    \label{tab:GD-VCR dataset}
\end{table}

\section{Full Performance}
\label{appendix:performance}

\begin{table*}[!ht]
\centering
\small
\begin{tabular}{lrrrrr}
\toprule
\textbf{Models}                   & \textbf{Overall} & \textbf{West}  & \textbf{South Asia} & \textbf{East Asia} & \textbf{Africa} \\ \midrule
\textit{Human Performance} & \textit{88.84}        &\textit{91.23}       &\textit{92.98}            & \textit{83.05}          & \textit{87.93} \\
\midrule
VisualBERT* & 53.27 & 62.91  & 52.04 & 45.39 & 51.85 \\
ViLBERT*  & 58.47 & 65.82 & 62.9  & 46.45 & 62.04 \\
\midrule
VL-BERT \textit{(seed 1)}    & 58.8 & 64.73 & 67.42 & 48.58 & 54.78 \\
VL-BERT \textit{(seed 2)}    & 58.92 & 65.82 & 63.8 & 47.16 & 62.04 \\
VL-BERT \textit{(seed 3)}    & 58.19 & 65.27 & 63.54 & 47.91 & 59.7 \\
\rowcolor{lightgray} VL-BERT \textit{(average)}  & 58.63 & 65.27 & 64.92 & 47.88 & 58.17 \\
\midrule
VLC-BERT with GD-BART \textit{(seed 1)} & 53.63 & 58.54 & 54.08 & 41.42 & 52.78 \\
VLC-BERT with GD-BART \textit{(seed 2)} & 52.92 & 57.27 & 54.08 & 42.09 & 52.13 \\
VLC-BERT with GD-BART \textit{(seed 3)} & 51.55 & 57.27 & 54.89 & 42.09 & 50.70 \\
\rowcolor{lightgray} VLC-BERT with GD-BART \textit{(average)}  & 52.69 & 57.69 & 54.35 & 41.87 & 51.87 \\

VLC-BERT with COMET \textit{(seed 1)} & 59.71 & 67.27 & 64.71 & 50.00 & 55.56 \\
VLC-BERT with COMET \textit{(seed 2)} & 59.82 & 67.27 & 64.71 & 50.36 & 55.56 \\
VLC-BERT with COMET \textit{(seed 3)} & 59.25 & 65.82 & 63.35 & 48.58 & 62.04 \\
\rowcolor{lightgray} VLC-BERT with COMET \textit{(average)}  & 59.59 & 66.78 & 64.25 & 49.64 & 57.71 \\
VLC-BERT with \model{} \textit{(seed1)} & 62.87 &  67.64 &70.14 &51.77 &64.81\\

VLC-BERT with \model{} \textit{(seed2)} & 65.01 & 72.36 & 67.42 &54.97 &67.59\\

VLC-BERT with \model{} \textit{(seed3)} & 62.64 & 69.82 &66.97 &52.48 &62.03\\

\rowcolor{lightgray} \textbf{VLC-BERT with \model{} \textit{(average)}} & \textbf{63.51} & \textbf{69.93} & \textbf{68.17} & \textbf{53.07} & \textbf{64.81}\\
\bottomrule
\end{tabular}
\caption{Accuracy (\%) of the various models on the subset of each region in GD-VCR. Results marked with $*$ have been reported in \newcite{yin-etal-2021-broaden}.}
\label{tab:full_results}
\end{table*}
Extending upon Table \ref{tab:results}, we provide a complete summary of the results of individual seeds on GD-VCR in Table \ref{tab:full_results}.

\end{document}